\documentclass[10pt,journal,compsoc]{IEEEtran}\pdfoutput=1
\usepackage{amsmath,amssymb,array,url,color,enumerate,amsfonts,booktabs,multirow,cite,subfigure,graphicx,bm}
\usepackage[bookmarksnumbered]{hyperref}
\newtheorem{dfn}{Definition}
\newcolumntype{L}[1]{>{\raggedright\let\newline\\\arraybackslash\hspace{0pt}}m{#1}}

\allowdisplaybreaks

\begin{document}
\title{A Survey on Negative Transfer}
\author{Wen~Zhang, Lingfei~Deng, Lei~Zhang,~\IEEEmembership{Senior~Member,~IEEE}, Dongrui~Wu,~\IEEEmembership{Senior~Member,~IEEE}
\IEEEcompsocitemizethanks{
\IEEEcompsocthanksitem Wen~Zhang, Lingfei~Deng and Dongrui~Wu are with the Key Laboratory of the Ministry of Education for Image Processing and Intelligent Control, School of Artificial Intelligence and Automation, Huazhong University of Science and Technology, Wuhan 430074, China. (e-mail: \{wenz,\ lfdeng,\ drwu\}@hust.edu.cn).
\IEEEcompsocthanksitem Lei~Zhang is with the School of Microelectronics and Communication Engineering, Chongqing University, Chongqing 400044, China (e-mail: leizhang@cqu.edu.cn).
\IEEEcompsocthanksitem Wen~Zhang and Lingfei~Deng contributed equally to this work. Dongrui~Wu is the corresponding author.}}

\IEEEtitleabstractindextext{
\begin{abstract}
Transfer learning (TL) utilizes data or knowledge from one or more source domains to facilitate the learning in a target domain. It is particularly useful when the target domain has very few or no labeled data, due to annotation expense, privacy concerns, etc. Unfortunately, the effectiveness of TL is not always guaranteed. Negative transfer (NT), i.e., leveraging source domain data/knowledge undesirably reduces the learning performance in the target domain, has been a long-standing and challenging problem in TL. Various approaches have been proposed in the literature to handle it. However, there does not exist a systematic survey on the formulation of NT, the factors leading to NT, and the algorithms that mitigate NT. This paper fills this gap, by first introducing the definition of NT and its factors, then reviewing about fifty representative approaches for overcoming NT, according to four categories: secure transfer, domain similarity estimation, distant transfer, and NT mitigation. NT in related fields, e.g., multi-task learning, lifelong learning, and adversarial attacks, are also discussed.
\end{abstract}

}
\maketitle

\IEEEraisesectionheading{\section{Introduction}}

\IEEEPARstart{A} common assumption in traditional machine learning is that the training data and the test data are drawn from the same distribution. However, this assumption does not hold in many real-world applications. For example, two image datasets may contain images taken using cameras with different resolutions under different light conditions; different people may demonstrate strong individual differences in brain-computer interfaces \cite{drwuTLBCI2021}. Therefore, the resulting machine learning model may generalize poorly.

A conventional approach to mitigate this problem is to re-collect a large amount of labeled or partly labeled data, which have the same distribution as the test data, and then train a machine learning model on the new data. However, many factors may prevent easy access to such data, e.g., high annotation cost, privacy concerns, etc.

A better solution to the above problem is transfer learning (TL) \cite{pan2009survey}, or domain adaptation (DA) \cite{zhang2019tdl}, which tries to utilize data or knowledge from related domains (called source domains) to facilitate the learning in a new domain (called target domain). TL was first studied in educational psychology to enhance human's ability to learn new tasks and to solve novel problems \cite{chen1989positive}. In machine learning, TL is mainly used to improve a model's ability to generalize in the target domain, which usually has zero or a very small number of labeled data. Many different TL approaches have been proposed, e.g., traditional (statistical) TL \cite{pan2011domain,long2013jda,dai2007boosting,zhang2017joint,drwuTHMS2017}, deep TL \cite{ghifary2014domain,long2015learning}, adversarial TL \cite{ganin2016domain,tang2020discriminative}, etc.

\begin{figure}[t]  \centering
\includegraphics[width=0.44\textwidth,clip]{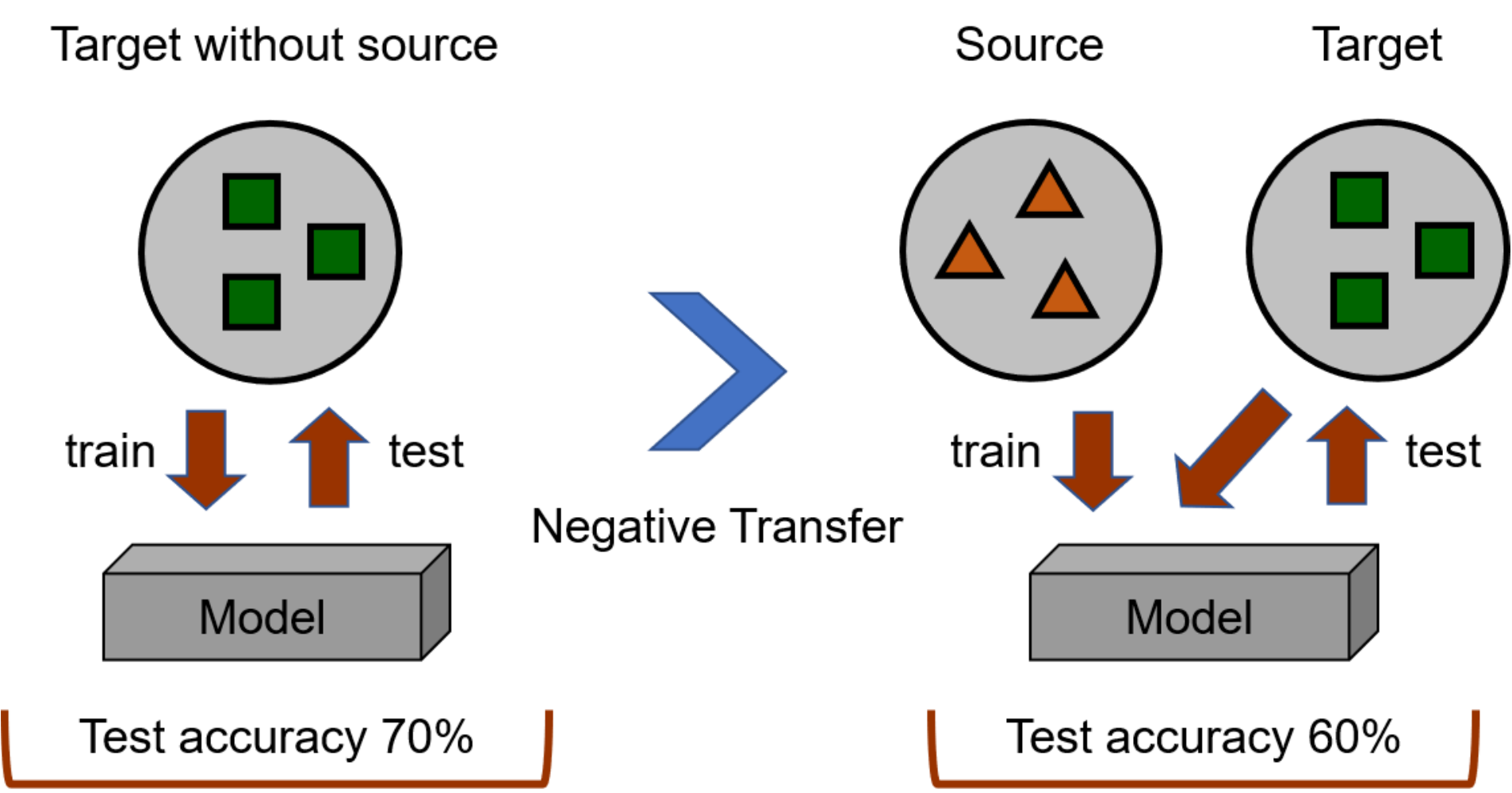}
\caption{Illustration of NT: introducing source domain data/knowledge decreases the target domain learning performance.} \label{fig:NT}
\end{figure}

Unfortunately, the effectiveness of TL is not always guaranteed, unless its basic assumptions are satisfied: 1) the learning tasks in the two domains are related/similar; 2) the source domain and target domain data distributions are not too different; and, 3) a suitable model can be applied to both domains. Violations of these assumptions may lead to \emph{negative transfer} (NT), i.e., introducing source domain data/knowledge undesirably decreases the learning performance in the target domain, as illustrated in Fig.~\ref{fig:NT}. NT is a long-standing and challenging problem in TL \cite{ros2005transfer,pan2009survey,wang2019characterizing}.

The following fundamental problems need to be addressed for reliable TL \cite{pan2009survey}: 1) what to transfer; 2) how to transfer; and, 3) when to transfer. Most TL research \cite{tan2018survey,zhang2019tdl} focused only on the first two, whereas all three should be taken into consideration to avoid NT. To our knowledge, NT was first studied in 2005 \cite{ros2005transfer}, and received increasing attention recently \cite{wang2019characterizing,li2019towards,leep2020}. Various ideas, e.g., finding similar parts of domains, evaluating the transferability of different tasks/models/features, etc., have been explored.

Though very important, there does not exist a comprehensive survey on NT. This paper aims to fill this gap, by systematically reviewing about fifty representative approaches to cope with NT. We mainly consider homogeneous and closed-set classification problems in TL, i.e., the source and target tasks are the same, and the target feature and label spaces are also unchanged during testing. This is the most studied TL scenario. We introduce the definition and factors of NT, methods that can avoid NT under theoretical guarantees, methods that can mitigate NT to a certain extent, and some related fields. Articles that do not explain their methods from the perspective of NT are not included in this survey, to keep it more focused.

The remainder of this paper is organized as follows. Section~2 introduces background knowledge in TL and NT. Section~3 proposes an scheme for reliable TL. Sections~4-7 review secure transfer, domain similarity estimation, distant transfer, and NT mitigation strategies, respectively. Section~8 introduces several related machine learning fields. Section~9 compares all reviewed approaches. Finally, Section~10 draws conclusions and points out some future research directions.

\section{Background Knowledge}

This section introduces some background knowledge on TL and NT, including the notations, definitions and categorizations of TL, and factors of NT.

\subsection{Notations and Definitions}

We consider classifiers with $K$ categories, with an input feature space $\mathcal{X}$ and an output label space $\mathcal{Y}$. Assume we have access to one labeled source domain $\mathcal{S}=\{(\bm{x}_s^i,y_s^i)\}_{i=1}^{n_s}$ drawn from $P_{\mathcal{S}}(X,Y)$, where $X\subseteq\mathcal{X}$ and $Y\subseteq\mathcal{Y}$. The target domain consists of two sub-datasets: $\mathcal{T} = (\mathcal{T}_l,\mathcal{T}_u)$, where $\mathcal{T}_l=\{(\bm{x}_l^j,y_l^j)\}_{j=1}^{n_l}$ consists of $n_l$ labeled samples drawn from $P_{\mathcal{T}}(X,Y)$, and $\mathcal{T}_u=\{\bm{x}_u^k\}_{k=1}^{n_u}$ consists of $n_u$ unlabeled samples drawn from $P_{\mathcal{T}}(X)$.
The main notations are summarized in Table~\ref{tab:notation}.

\begin{table}[htbp] \centering
  \caption{Main notations in this survey.}
    \begin{tabular}{ll|ll} \toprule
    \multicolumn{1}{l}{Notation} & \multicolumn{1}{l|}{Description} & \multicolumn{1}{l}{Notation} & \multicolumn{1}{l}{Description} \\ \midrule
        $\bm{x}$  & Feature vector  &  $\ell(\cdot), \mathcal{L}(\cdot)$ &  Loss function \\
        $y$  & Label of $\bm{x}$  &  $h$    & Hypothesis \\
        $\mathcal{X}$  &   Feature space    &$f$    & Classifier \\
        $\mathcal{Y}$  &   Label space    &    $\theta$    & TL algorithm\\
        $\mathcal{S}$  &   Source domain    &$g$  &   Feature extractor \\
        $\mathcal{T}$  &   Target domain    & $\epsilon$    & Error (risk) \\
        $P(\cdot)$  &   Distribution  &$n$   & Number of samples \\
        $\mathbb{E}(\cdot)$ &  Expectation  &$K$   & Number of classes \\
         $d(\cdot)$   & Distance metric    &   $M$  & No. of source domains \\ \bottomrule
    \end{tabular}
  \label{tab:notation}
\end{table}

In TL, the condition that the source and target domains are different (i.e., $\mathcal{S} \neq \mathcal{T}$) implies one or more of the following:
\begin{enumerate}
\item The feature spaces are different, i.e., $\mathcal{X}_{\mathcal{S}} \neq \mathcal{X}_{\mathcal{T}}$.
\item The label spaces are different, i.e., $\mathcal{Y}_{\mathcal{S}} \neq \mathcal{Y}_{\mathcal{T}}$.
\item The marginal probability distributions of the two domains are different, i.e., $P_{\mathcal{S}}(X) \neq P_{\mathcal{T}}(X)$.
\item The conditional probability distributions of the two domains are different, i.e., $P_{\mathcal{S}}(Y | X) \neq P_{\mathcal{T}}(Y | X)$.
\end{enumerate}
This survey focuses on the last two differences, i.e., we assume that the source and target domains share the same feature and label spaces. This is actually the case of DA.

TL aims to design a learning algorithm $\theta(\mathcal{S},\mathcal{T})$, which utilizes data/information in the source and target domains to output a hypothesis $h$ as the target domain mapping function, with a small expected loss $\epsilon_\mathcal{T}(h)=\mathbb{E}_{\bm{x},y\sim P_{\mathcal{T}}(X,Y)}[\ell(h(\bm{x}),y)]$, where $\ell$ is a target domain loss function.

\subsection{TL Categorization}

According to \cite{pan2009survey}, TL approaches can be categorized into four groups: instance based, feature based, model/parameter based, and relation based.

Instance based approaches mainly focus on sample weighting, assuming the distribution discrepancy between the source and target domains is caused by a sample selection bias, which can be compensated by reusing a certain portion of the weighted source domain data \cite{zhang2019tdl,zhang2017joint}.

Feature based approaches aim to find a latent subspace or representation to match the two domains, assuming there exists a common space in which distribution discrepancies of different domains can be minimized \cite{zhang2019tdl}.

Model/parameter based approaches transfer knowledge via parameters, assuming the distributions of model parameters in different domains are the same or similar \cite{agrawal2014fine,liang2020shot}.

Relation based approaches assume that some internal logical relationships or rules in the source domain are preserved in the target domain.

More details on TL can be found in \cite{pan2009survey,zhang2019tdl,yang2020transfer,zhuang2020survey}.

\subsection{NT} \label{sect:form}

Rosenstein \emph{et al.} \cite{ros2005transfer} first discovered NT through experiments, and concluded that ``\emph{transfer learning may actually hinder performance if the tasks are too dissimilar}" and ``\emph{inductive bias learned from the auxiliary tasks will actually hurt performance on the target task.}" Pan \emph{et al.} \cite{pan2009survey} also briefly mentioned NT in their TL survey: ``\emph{When the source domain and target domain are not related to each other, brute-force transfer may be unsuccessful. In the worst case, it may even hurt the performance of learning in the target domain, a situation which is often referred to as negative transfer.}"

Wang \emph{et al.} \cite{wang2019characterizing} gave a mathematical definition of NT, and proposed a negative transfer gap (NTG) to determine whether NT happens.
\begin{dfn}
(Negative transfer gap \cite{wang2019characterizing}). Let $\epsilon_{\mathcal{T}}$ be the test error in the target domain, $\theta(\mathcal{S}, \mathcal{T})$ a TL algorithm between $\mathcal{S}$ and $\mathcal{T}$, and $\theta(\emptyset, \mathcal{T})$ the same algorithm but does not use the source domain information at all. Then, NT happens when $\epsilon_{\mathcal{T}} \left(\theta(\mathcal{S}, \mathcal{T})\right) > \epsilon_{\mathcal{T}} \left(\theta(\emptyset, \mathcal{T})\right)$, and the degree of NT can be evaluated by the NTG:
\begin{align}
NTG = \epsilon_{\mathcal{T}} (\theta(\mathcal{S}, \mathcal{T})) - \epsilon_{\mathcal{T}} (\theta(\emptyset, \mathcal{T})). \label{eq:ntg} \hfill \blacksquare
\end{align}
\end{dfn}

Obviously, NT occurs if the NTG is positive. However, NTG may not always be computable. For example, in an unsupervised scenario, $\epsilon_{\mathcal{T}} (\theta(\emptyset, \mathcal{T}))$ is impossible to compute due to the lack of labeled target data.

\subsection{Factors of NT} \label{sect:factor}

Ben-David \emph{et al.} \cite{ben2010theory} gave a theoretical bound for TL:
\begin{align}
\epsilon_{\mathcal{T}}(h)\le \epsilon_{\mathcal{S}}(h)+\frac{1}{2}d_{\mathcal{H}\Delta \mathcal{H}}(X_s,X_t)+\lambda, \label{eq:tlbound}
\end{align}
where $\epsilon_{\mathcal{T}}(h)$ and $\epsilon_{\mathcal{S}}(h)$ are respectively the expected error of hypothesis $h$ in the source domain and the target domain, $d_{\mathcal{H}\Delta \mathcal{H}}(X_s,X_t)$ is the domain divergence between the two domains, and $\lambda$ is a problem-specific constant.

Based on (\ref{eq:tlbound}), the following four factors could contribute to NT:
\begin{enumerate}
\item \emph{Domain divergence.} Arguably, the divergence between the source and target domains is the root of NT. TL approaches that do not explicitly consider minimizing the divergence, whether at the feature, classifier, or target output level, are more likely to result in NT.
\item \emph{Transfer algorithm.} A secure transfer algorithm should have a theoretical guarantee that the learning performance in the target domain is better when auxiliary data are utilized, or the algorithm has been carefully designed to improve the transferability of auxiliary domains. Otherwise, NT may happen.
\item \emph{Source data quality.} Source data quality determines the quality of the transferred knowledge. If the source data are inseparable or very noisy, then a classifier trained on them may be unreliable. Sometimes the source data have been converted into pre-trained models, e.g., for privacy-preserving. An over-fitting or under-fitting source domain model may also cause NT.
\item \emph{Target data quality.} The target domain data may be noisy and/or non-stationary, which may also lead to NT.
\end{enumerate}

\section{Reliable TL}

\begin{figure}[htpb] \centering
\includegraphics[width=\linewidth,clip]{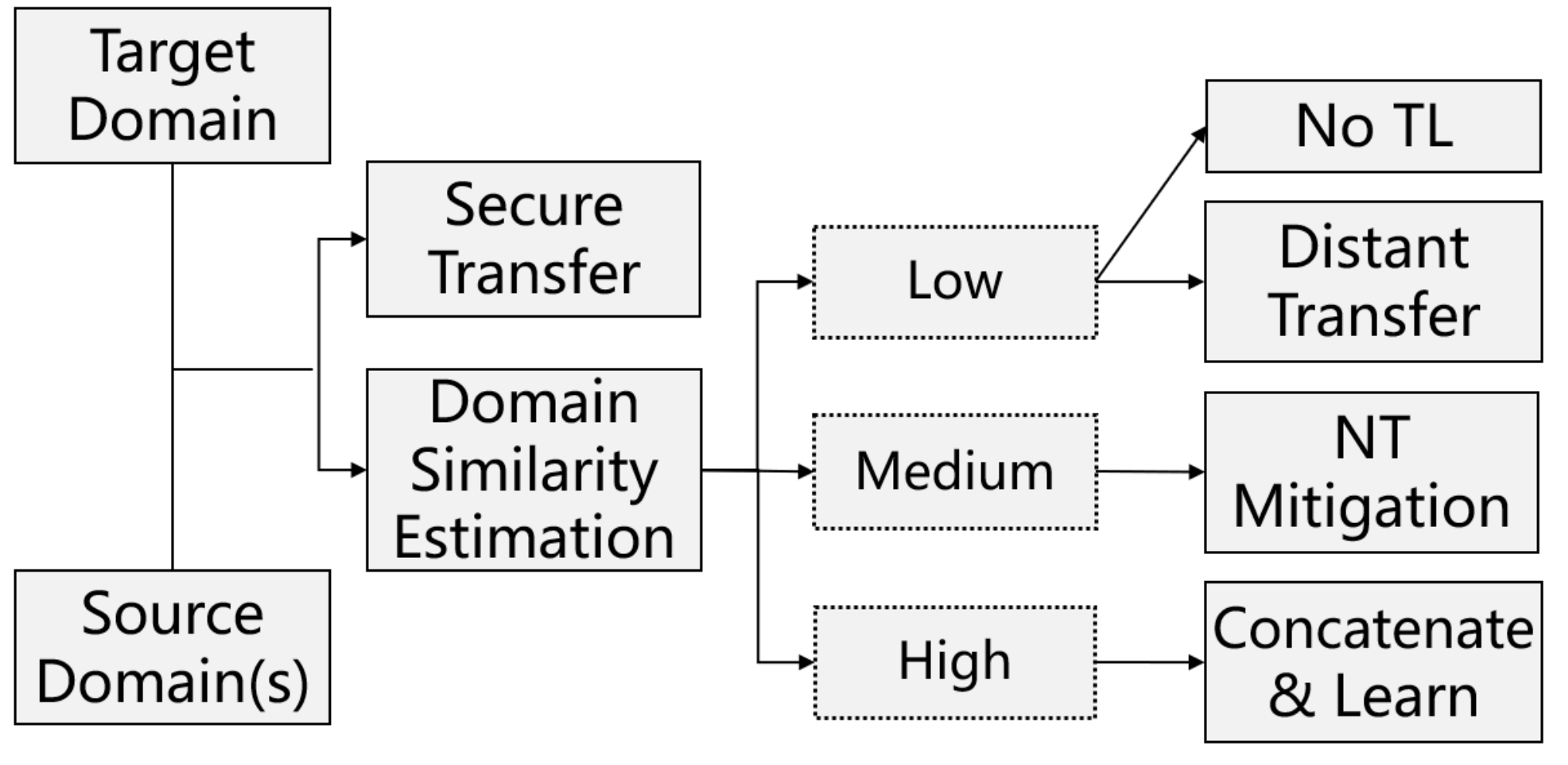}
\caption{Our proposed reliable TL scheme.} \label{fig:overview}
\end{figure}

Fig.~\ref{fig:overview} shows our proposed reliable TL scheme, considering existing typical strategies for alleviating or avoiding NT.

Secure transfer can overcome NT with theoretical guarantees, regardless of whether the source and target domains are similar or not. Most other approaches assume the source domain has some similarity with the target domain. Accurately estimating this similarity helps determine which strategy should be used to handle NT.

With the estimated domain similarity, we can: 1) refuse to transfer or use distant transfer when the similarity is low; 2) perform NT mitigation when the similarity is medium; or, 3) concatenate data from different domains directly and train a classifier from them, when the similarity is high.

When there are multiple source domains, domain similarity estimation can also be used to select the most transferable source domains.

\section{Secure Transfer}

Secure transfer explicitly avoids NT in its objective function, i.e., the TL algorithm should perform better than the one without transfer, which is appealing in real-world applications, especially when it is difficult to estimate the domain similarity.

As shown in Table~\ref{tab:secure}, there are only a few secure transfer approaches. Some of them consider only classification problems, whereas others consider only regression problems.

\begin{table}[htbp] \centering
  \caption{Existing secure transfer approaches.}
    \begin{tabular}{p{11em}p{10em}p{5em}} \toprule
    \textsc{Category}          & \textsc{Approach}  & \textsc{Reference}   \\ \midrule

    \textbf{Classification Transfer}   & Adaptive learning    & \cite{cao2010adaptive} \\
                               & Performance gain    & \cite{abdullah2018deep,li2019towards} \\ \midrule

    \textbf{Regression Transfer}   & Output truncation   & \cite{kuzborskij2013stability} \\
                               & Regularization    & \cite{yoon2018novel} \\
                               & Bayesian optimization    & \cite{Sorocky2021} \\ \bottomrule
    \end{tabular}
   \label{tab:secure}
\end{table}

\subsection{Secure Transfer for Classification}

Cao \emph{et al.} \cite{cao2010adaptive} proposed a Bayesian adaptive learning approach to adjust the transfer schema automatically according to the similarity of the two tasks. It assumes the source and target data obey a Gaussian distribution with a semi-parametric transfer kernel $\bm{K}$,
\begin{align}
\bm{K}_{nm} \sim k(\bm{x}_n, \bm{x}_m)(2e^{-\varsigma(\bm{x}_n, \bm{x}_m)\rho} - 1),
\end{align}
where $k$ is a valid kernel function. $\varsigma(\bm{x}_n, \bm{x}_m)=0$ if $\bm{x}_n$ and $\bm{x}_m$ are from the same domain, and $\varsigma(\bm{x}_n, \bm{x}_m)=1$ otherwise. The parameter $\rho$ represents the dissimilarity between the source and target domains. By assuming $\rho$ is from a Gamma distribution $\Gamma(b, \mu)$, where $b$ and $\mu$ are respectively the shape and the scale parameters inferred from a few labeled samples in both domains, we can define
\begin{align}
\lambda=2 \left(\frac{1}{1+\mu} \right)^b - 1,
\end{align}
which determines the similarity between the two domains, and also what can be transferred. For example, when $\lambda$ is close to $0$, the correlation between the domains is low, so only the parameters in the kernel function $k$ can be shared.

Jamal \emph{et al.} \cite{abdullah2018deep} proposed a deep face detector adaptation approach to avoid NT and catastrophic forgetting in deep TL, by minimizing the following loss function:
\begin{align}
\min_{\bm{u},\tilde{\bm{\theta}}}\ \left[\frac {\lambda} 2 \left\|\bm{u}\right\|_2^2 + \mathbb{E}_t\max_{y_t\in\{0,1\}}\ RES_t\left(\bm{w}+\bm{u},\tilde{\bm{\theta}}\right)\right],
\end{align}
where $\bm{w}$ are the classifier weights of the source detector, $\bm{u}$ the offset weights to constrain the target face detector around the source detector, and $\tilde{\bm{\theta}}$ the parameters of the target feature extractor. $RES_t$ is the relative performance loss of the learned target detector over the pre-trained source face detector, which is non-positive after optimization. Hence, the obtained target detector is always no worse than the source detector, i.e., NT is avoided.

Li \emph{et al.} \cite{li2019towards} proposed a safe weakly supervised learning (SAFEW) approach for semi-supervised DA. Assume the target hypothesis $h^{\ast}$ can be constructed from multiple base learners in the source domains, i.e., $h^{\ast} = \sum_{i=1}^M\alpha_i h_i$, where $\{h_i\}_{i=1}^M$ are the $M$ source models with $\bm{\alpha}=[\alpha_1;\alpha_2;...;\alpha_M]\ge \bm{0} $ and $\sum_{i=1}^M \alpha_i =1$. The goal is to learn a prediction $h$ that maximizes the performance gain against the baseline $h_0$, which is trained on the labeled target data only, by optimizing the following objective function:
\begin{align}
\max_{h}\min_{\bm{\alpha}\in \mathcal{M}}\ \ell\left(h_0,\sum_{i=1}^M \alpha_i h_i\right) - \ell\left(h,\sum_{i=1}^M \alpha_i h_i\right),
\end{align}
i.e., SAFEW optimizes the worst-case performance gain to avoid NT.

\subsection{Secure Transfer for Regression}

Kuzborskij and Orabona \cite{kuzborskij2013stability} introduced a class of regularized least squares (RLS) \cite{bishop2006pattern} algorithms with biased regularization to avoid NT. The original RLS algorithm solves the following optimization problem:
\begin{align}
\min_{\bm{w}}\ \frac 1 n \sum_{i=1}^n\left(\bm{w}^{\top}\bm{x}_i-y_i\right)^2 + \lambda \|\bm{w}\|^2.
\end{align}
After obtaining the optimized source hypothesis $h'(\cdot)$, the authors constructed a training set  $\{(\bm{x}_i,y_i-h'(\bm{x}_i))\}_{i=1}^n$, and generated the transfer hypothesis
\begin{align}
h_{\mathcal{T}}(\bm{x})=T_C(\bm{x}^{\top} \hat{\bm{w}}_{\mathcal{T}}) + h'(\bm{x}),
\end{align}
where $T_C(\hat{y})=\min(\max(\hat{y},-C),C)$ is a truncation function to limit the output to $[-C,C]$, and
\begin{align}
\hat{\bm{w}}_{\mathcal{T}}=\arg \min\limits_{\bm{w}} \frac 1 n \sum_{i=1}^n(\bm{w}^{\top}\bm{x}_i-y_i+h'(\bm{x}_i))^2 + \lambda \|\bm{w}\|^2.
\end{align}
Kuzborskij and Orabona \cite{kuzborskij2013stability} showed that the proposed approach is equivalent to RLS trained solely in the target domain when the source domains are unrelated to the target domain.

Yoon and Li \cite{yoon2018novel} proposed a positive TL approach, based also on the RLS algorithm. It assumes the source parameters follow a normal distribution, and optimizes the following loss function:
\begin{align}
\min_{\bm{w}}\ \ell_{\mathcal{T}_l}(\bm{w};b) + \beta \mathcal{R}(\bm{w}) + \lambda N(\bm{w}; \bm{\mu}_{\bm{w}}, \Sigma_{\bm{w}}),
\end{align}
where $\bm{w}$ are model coefficients, $\mathcal{R}(\bm{w})$ a regularization term to control the model complexity, and $N(\bm{w}; \bm{\mu}_{\bm{w}}, \Sigma_{\bm{w}})$ a regularization term to constrain $\bm{w}$ in a multi-variable Gaussian distribution with mean $\bm{\mu}_{\bm{w}}$ and variance $\Sigma_{\bm{w}}$ computed from the source domains. They showed that NT arises when $\lambda$ is too large, thus proposed an optimization rule to select the weight $\lambda$ and hence to eliminate harmful source domains.

Sorocky \emph{et al.} \cite{Sorocky2021} derived a theoretical bound on the test error and proposed a Bayesian-optimization based approach to estimate this bound to guarantee positive transfer in a robot tracking system. Firstly, they bounded the 2-norm of the tracking error of the target robot using the source module by
\begin{align}
\left \| e_{t,s}\right \|_{2} \leq \left \| E_{t,s}\right \|_{\infty} \left \| y_d \right \|_{2},
\end{align}
where $E_{t,s}$ represents the transfer function of the robot tracking system, and $y_d$ the desired output of the source module. Given the baseline target tracking error $e_{t,b}$, this bound can guarantee positive transfer if $\left \| e_{t,b}\right \|_{2} \geq \left \| E_{t,s}\right \|_{\infty} \left \| y_d \right \|_{2}$. Since $y_d$ is fixed and known, the authors established a Gaussian Process model to estimate $\left \| E_{t,s}\right \|_{\infty}$ and compute the error bound, guaranteeing positive transfer.

\section{Domain Similarity Estimation} \label{sect:dse}

Domain similarity (or transferability) estimation is a very important block in reliable TL, as shown in Fig.~\ref{fig:overview}. Existing estimation approaches can be categorized into three groups: feature statistics based, test performance based, and fine-tuning based, as summarized in Table~\ref{tab:similarity}.


\begin{table}[htbp] \centering
  \caption{Approaches for domain similarity estimation.}
    \begin{tabular}{p{9em}p{9em}p{8em}} \toprule
    \textsc{Category}           & \textsc{Approach}  & \textsc{Reference}   \\ \midrule

    \textbf{Feature Statistics Based} & MMD            & \cite{gretton2012kernel} \\
                                & Correlation    & \cite{lin2017improving,zhang2020manifold} \\
                                & KL divergence  & \cite{Gong2012,azab2019weighted} \\ \midrule

    \textbf{Test Performance Based}   & Target performance       & \cite{yao2010boosting,xie2017selective} \\
                                & Domain classifier    & \cite{ben2007analysis,xu2018deep,wu2020continuous} \\ \midrule

    \textbf{Fine-Tuning based}        & Clustering quality      & \cite{Meiseles2020} \\
                                & Entropy      & \cite{tran2019transfer,afridi2018automated,leep2020,huang2021frustratingly} \\ \bottomrule
    \end{tabular}
   \label{tab:similarity}
\end{table}

\subsection{Feature Statistics Based}

The original feature representation and its first- or high- order statistics, such as mean and covariance, are direct and important inputs for measuring the domain distribution discrepancy. Three widely used domain discrepancy measurements are maximum mean discrepancy (MMD), correlation coefficient, and KL-divergence.

MMD \cite{gretton2012kernel} may be the most popular discrepancy measure in traditional TL \cite{pan2011domain,long2013jda,zhang2017joint,wenz20djpmmd}, due to its simplicity and effectiveness. It is a nonparametric measure, and can be computed directly from the feature means in the raw feature space or a mapped Reproducing Kernel Hilbert Space (RKHS). Empirically, the MMD between the source and target domains can be computed via:
\begin{align}
\begin{split}
MMD^2(\mathcal{S}, \mathcal{T}) &= \left\|\frac{1}{n_{s}} \sum_{i=1}^{n_{s}} \bm{x}_{s, i}-\frac{1}{n_{t}} \sum_{j=1}^{n_{t}} \bm{x}_{t, j} \right\|_2^{2},
\end{split}
\end{align}
where $k$ is a kernel function, and $\bm{x}_s$ and $\bm{x}_t$ are source and target domain samples, respectively. However, MMD only considers the marginal distribution discrepancy between different domains, so it may fail when the conditional distributions between the source and target domains are also significantly different.

The correlation between two high-dimensional random variables from different distributions can also be used to evaluate the distribution discrepancy. Lin and Jung \cite{lin2017improving} evaluated the inter-subject similarity in emotion classification via the correlation coefficient of the original feature representations from two different subjects.

To fully utilize the source label information, Zhang and Wu \cite{zhang2020manifold} developed a domain transferability estimation (DTE) index to evaluate the transferability between a source domain and a target domain via between-class and between-domain scatter matrices:
\begin{align}
DTE(\mathcal{S},\mathcal{T}) = \frac {\|S_b^{\mathcal{S}}\|_1} {\|S_b^{\mathcal{S},\mathcal{T}}\|_1},
\end{align}
where $S_b^{\mathcal{S}}$ is the between-class scatter matrix in the source domain, $S_b^{\mathcal{S},\mathcal{T}}$ the between-domain scatter matrix, and $\|\cdot\|_1$ the 1-norm. DTE has low computational cost and is insensitive to the sample size.

KL-divergence \cite{kullback1951kl} is a non-symmetric measure of the divergence between two probability distributions. Gong \emph{et al.} \cite{Gong2012} proposed a rank of domain (ROD) approach to rank the similarities of the source domains to the target domain, by computing the symmetrized KL divergence weighted average of the principal angles. It can be used to automatically select the optimal source domains to adapt. Azab \emph{et al.} \cite{azab2019weighted} computed the similarity $\alpha_s$ between the target domain feature set $d_t$ and the source domain feature set $d_s$ as:
\begin{align}
\alpha_s=\frac{1/\left(\overline{KL}[d_t,d_s]+\epsilon\right )^4}{\sum_{m=1}^M\left (1/\left (\overline{KL}[d_t,d_m]+\epsilon\right)^{4}\right )},
\end{align}
where $\overline{KL}$ represents the average per-class KL-divergence, $M$ the number of source domains, and $\epsilon=0.0001$ is used to ensure the stability of calculation.

Additionally, Hilbert-Schmidt independence criterion (HSIC) \cite{gretton2005measuring}, Bregman divergence \cite{si2009bregman}, optimal transport and  Wasserstein distance \cite{shen2018wasserstein}, etc., have also been used to measure the feature distribution discrepancy in conventional TL.

\subsection{Test Performance Based}

The domain similarity can also be measured from the test performance: if a source domain classifier can achieve good performance on the labeled target domain data, then that source domain and the target domain should be very similar. However, this requires the target domain to have some labeled data.

Yao and Doretto \cite{yao2010boosting} used $M$ iterations to train $M$ weak classifiers from $N$ source domains ($M\le N$) and then combined them as the final classifier. In the $m$-th iteration, the classifier with the smallest error on the labeled target domain samples was chosen as the $m$-th weak classifier.

Some works assume the target model is accessible for source selection. Xie \emph{et al.} \cite{xie2017selective} proposed selective transfer incremental learning (STIL) to remove less relevant source (historical) models for online TL. STIL computes the following Q-statistic as the correlation between a historical model and the newly trained target model:
\begin{align}
Q_{f_i,f_j} = \frac {N^{11}N^{00} - N^{01}N^{10}} {N^{11}N^{00} + N^{01}N^{10}},
\end{align}
where $f_i$ and $f_j$ are two classifiers. $N^{y_iy_j}$ is the number of instances for which the classification result is $y_i$ by $f_i$ ($y_i=1$ if $f_i$ classifies the example correctly; otherwise $y_i=0$), and $y_j$ by $f_j$. STIL then removes the less transferable historical models, whose Q-statistics are close to $0$. In this way, it can overcome NT. This strategy was also used in \cite{sun2018concept}.

The above approaches require to know a sufficient number of labeled target samples, or the target model, which may not always be available. In such cases, discriminator based similarity measures could be used. These approaches train classifiers to discriminate the two domains and then define a similarity measure from the classification error \cite{ben2010theory, xu2018deep}.

Ben-David \emph{et al.} \cite{ben2007analysis} proposed an unsupervised $\mathcal{A}$-distance to find the minimum-error classifier
\begin{align}
d_{\mathcal{A}}(\bm{\mu}_S,\bm{\mu}_T) = 2\left(1-2\min_{h\in \mathcal{H}}\epsilon(h)\right),
\end{align}
where $\mathcal{H}$ is the hypothesis space, $h$ a domain classifier, and $\epsilon(h)$ the domain classification error. The $\mathcal{A}$-distance should be small for good transferability. Unfortunately, computing $d_{\mathcal{A}}(\bm{\mu}_S,\bm{\mu}_T)$ is NP-hard. To reduce the computational cost, they trained a linear classifier to determine which domain the data come from, and utilized its error to approximate the optimal classifier. However, $\mathcal{A}$-distance neglects the difference in label spaces and only considers the marginal distribution discrepancy between domains, which may affect its performance.

Recently, Wu and He \cite{wu2020continuous} proposed a novel label-informed divergence between the source and target domains, when the target domain is time evolving. This divergence can measure the shift of joint distributions, which improves the $\mathcal{A}$-distance.

\subsection{Fine-Tuning Based}

The domain similarity can also be estimated from fine-tuning \cite{donahue2014decaf,agrawal2014analyzing}, which is frequently used in deep TL to adapt a source domain deep learning model to the target domain, by fixing its lower layer parameters and re-tuning the higher layer parameters.

Generally, these approaches feed the target data into the source neural network, and use the output to determine the domain similarity. Meiseles \emph{et al.} \cite{Meiseles2020} introduced a clustering quality metric, mean silhouette coefficient \cite{rousseeuw1987msc}, to assess the quality of the target encodings produced by a given source model. They found that this metric has the potential for source model selection. Tran \emph{et al.} \cite{tran2019transfer} developed negative conditional entropy (NCE), which measures the amount of information from a source domain to the target domain, to evaluate the source domain transferability.

Recently, computationally-efficient domain similarity measures without the source data have attracted great attention \cite{leep2020,afridi2018automated,huang2021frustratingly}. Nguyen \emph{et al.} \cite{leep2020} proposed log expected empirical prediction (LEEP), which can be computed from a source model $\theta$ with $n_l$ labeled target data, by running the target data through the model only once:
\begin{align}
T(\theta,\mathcal{D}) = \frac {1}{n_l} \sum_{i=1}^{n_l} \log\left(\sum_{z\in Z}
\hat{P}(y_i|z)\theta(x_i)_z\right), \label{eq:leep}
\end{align}
where $\hat{P}(y_i|z)$ is the empirical conditional distribution of the real target label $y_i$ given the dummy target label $z$ predicted by model $\theta$. $T(\theta,\mathcal{D})$ represents the transferability of the pre-trained model $\theta$ to the target domain $D$, and is an upper bound of the NCE measure.

In addition, Huang \emph{et al.} \cite{huang2021frustratingly} developed transfer rate (TransRate) for transferability estimation in fine-tuning based TL. Its motivation comes from mutual information of the output from the pre-trained feature extractor:
\begin{align}
\text{TrR}_{\mathcal{S}\to \mathcal{T}}(g,\epsilon) = R(Z, \epsilon) - R(Z, \epsilon|Y), \label{eq:transrate}
\end{align}
where $Y$ are labels of the target instances, $Z=g(X)$ features extracted by the pre-trained feature extractor $g$, and $R(Z, \epsilon)$ the rate distortion of $H(Z)$ to encode $Z$ with an expected decoding error less than $\epsilon$. They showed that TransRate has superior performance in selecting the source data, source model architecture, and even network layers.

\section{Distant Transfer}

NT may happen easily when the source and target data have very low similarity. For example, using text data as the auxiliary source is likely to reduce the performance of the target classifier originally trained on image data, causing NT. Distant TL (also called transitive TL) \cite{tan2015transitive}, which bridges dramatically different source and target domains through one or more intermediate domains, could be a solution to this problem.

Tan \emph{et al.} \cite{tan2017distant} introduced an instance selection mechanism to identify useful source data, and constructed multiple intermediate domains. They learned a pair of encoding function $f_e(\cdot)$ and decoding function $f_d(\cdot)$ to minimize the reconstruction errors on the selected instances in the intermediate domains, and also on all instances in the target domain simultaneously:
\begin{align}
\begin{split}
\mathcal{L}(f_e,f_d,v_{S},v_{I})=
R(v_{S},v_I)+
\frac{1}{n_{S}}\sum_{i=1}^{n_s}v_{S}^{i}\|\hat{\bm{x}}_{S}^{i}-\bm{x}_{S}^{i}\|_{2}^{2} \\
\frac{1}{n_I}\sum_{i=1}^{n_I}v_{I}^{i}\|\hat{\bm{x}}_{I}^{i}-\bm{x}_{I}^{i}\|_{2}^{2}+
\frac{1}{n_{T}}\sum_{i=1}^{n_t}\|\hat{\bm{x}}_{T}^{i}-\bm{x}_{T}^{i}\|_{2}^{2},
\end{split}
\end{align}
where $\hat{\bm{x}}_{S}^{i}$, $\hat{\bm{x}}_{T}^{i}$ and $\hat{\bm{x}}_{I}^{i}$ are reconstructions of $\bm{x}_{S}^{i}$, $\bm{x}_{T}^{i}$ and $\bm{x}_{I}^{i}$ from an auto-encoder, $v_{S}$ and $v_I$ are selection indicators, and $R(v_{S},v_I)$ is a regularization term. They also incorporated side information, such as predictions in the intermediate domains, to help the model learn more task-related feature representations.

Similar strategies have also been used in applications where the training data are very scarce, such as medical diagnostics and remote sensing. Niu \emph{et al.} \cite{niu2020distant} proposed a distant domain TL method that transfers knowledge from object recognition datasets, chest X-ray images, etc., to coronavirus diagnosis. They developed a convolutional auto-encoder pair to reconstruct both common image domains and medical image domains in the same intermediate feature space. All task-related information that may induce NT was removed after reconstruction. Xie \emph{et al.} \cite{Xie2016transfer} proposed a feature-based method to enrich scarce daytime satellite image data, by transferring knowledge learned from an object classification task, using the night-time light intensity information as a bridge.

\section{NT Mitigation}

In most TL problems, the source and target domains have some similarity, which could be used to mitigate NT.
This can be achieved by data transferability enhancement, model transferability enhancement, and target prediction enhancement, as shown in Table~\ref{tab:mitigation}.

\begin{table}[htbp] \centering
  \caption{Approaches for NT mitigation.}
    \begin{tabular}{p{10em}p{9em}p{7em}} \toprule
    \textsc{Category}             & \textsc{Approaches}  & \textsc{References}   \\ \midrule

   \textbf{Data Transferability}  & Domain level         & \cite{wang2018towards,zuo2021attention,ahmed2021unsupervised} \\
   \textbf{Enhancement}           & Instance level       & \cite{seah2012combating,drwuTNSRE2016,xu2020multi,peng20atl,wang2019characterizing} \\
                                  & Feature level        &
                                  \cite{long2012dual,shi2013twin,xu2015discriminative,rajesh2017annoyed,chen2019catastrophic,chen2019transfer,chen2020harmonizing}\\
                                  \midrule

   \textbf{Model Transferability} & TransNorm            & \cite{wang2019transferable} \\
   \textbf{Enhancement}           & Adversarial robust   & \cite{liang2020does,salman2020adversarially,deng2021adversarial} \\ \midrule
   \textbf{Target Prediction}    & Soft labeling        & \cite{pei2018multi,ge2020mutual} \\
   \textbf{Enhancement}           & Selective labeling      & \cite{gui2018negative,wang2020spl} \\
                                  & Weighted clustering    & \cite{liang2019exploring,wang2020spl,liang2020shot} \\ \bottomrule
    \end{tabular}
   \label{tab:mitigation}
\end{table}

\subsection{Data Transferability Enhancement}

The transferability of the source domain can be enhanced by improving the data quality from coarse to fine-grained, at the domain level, instance level, or feature level.

At the domain level, when there are multiple source domains, we can select a subset of them or weight them. At the instance level, we can select or weight the source instances. At the feature level, we can transform the original features into a common latent space or enhance their transferability.

\subsubsection{Domain Level Transferability Enhancement}

When there are multiple source domains, selectively aggregating the most similar ones to the target domain, or a weighted aggregation of all source domains, may achieve better performance than simply averaging all source domains \cite{eaton2011selective,yu2012kmm}. Therefore, domain selection/weighting can be used to mitigate NT.

The approaches introduced in Section~\ref{sect:dse} for estimating the similarity between a single source domain and a single target domain can be easily extended to multi-source TL scenarios. For example, Wang and Carbonell \cite{wang2018towards} used MMD to measure the proximity between the source and the target domains. They first trained a classifier in each source domain and then weighted them for target domain prediction, where each weight was a combination of the source domain's MMD-based proximity to the target domain and its transferability to other source domains. In the special case that the confidence of a source domain classifier is low, its own classification is discarded; instead, it queries its peers on this specific test sample, and each peer is weighted by its transferability to the current source domain.

The domain similarity (weight) can also be obtained through optimization \cite{zuo2021attention,ahmed2021unsupervised}. Zuo \emph{et al.} \cite{zuo2021attention} introduced an attention-based domain recognition module to estimate the domain correlations to alleviate the effects caused by dissimilar domains. Its main idea is to reorganize the instance labels when there are multiple source domains so that it can simultaneously distinguish each category and domain. It redefines the source labels by $\hat{Y}_{s,i}={Y}_{s,i} + (i-1)\times K$, and trains a domain recognition model on the original features. The learned weight of the $i$-th domain is
\begin{align}
w_i=\frac {\sum_{j=1}^{n_t} \mathrm{sign}(\hat{d}_j,i)} {n_t},
\end{align}
where $n_t$ is the target image number in a batch, $\mathrm{sign}(\cdot, \cdot)$ a sign function, and $\hat{d}_j$ the domain label of a target instance $x_j$ by analyzing the domain recognition model prediction. The authors verified that the learned domain weights had a high correlation with the groundtruth.

Ahmed \emph{et al.} \cite{ahmed2021unsupervised} considered optimizing the weights in a more challenging scenario when the source domain data are absent, and only the pre-trained source models are accessible. They first developed a sophisticated loss function $\mathcal{L}_{tar}$ from the source model predictions on the unlabeled target data. Then, the domain weights were optimized by,
\begin{align}
\begin{split}
\min_{\{\alpha_i\}_{i=1}^M}\ \mathcal{L}_{tar},\quad
\text{s.t.}\ \ \sum_{i=1}^M \alpha_i=1, \alpha_i\ge 0.
\end{split}
\end{align}
With the learned weights, the target model performs at least as good as the single best source model.

\subsubsection{Instance Level Transferability Enhancement}

Instance selection/weighting are frequently used in TL \cite{pan2009survey,zhang2019tdl,zhuang2020survey}. They can also be used to mitigate NT.

Seah \emph{et al.} \cite{seah2012combating} proposed a predictive distribution matching (PDM) regularizer to remove irrelevant source domain data. It iteratively infers the pseudo-labels of the unlabeled target domain data and retains the highly confident ones. Finally, an SVM or logistic regression classifier is trained using the remaining source domain data and the pseudo-labeled target domain data. Yi \emph{et al.} \cite{xu2020multi} partitioned source data into components by clustering, and assigned them different weights by iterative optimization.

Active learning \cite{drwuSAL2019,drwuiGS2019}, which selects the most useful unlabeled samples for labeling, can also be used to select the most appropriate source samples \cite{peng20atl, peng2020non}. Peng \emph{et al.} \cite{peng20atl} proposed active TL to optimally select source samples that are class balanced and highly similar to those in the target domain. It simultaneously minimizes the MMD and mitigates NT. If the unlabeled target domain samples can be queried for their labels, then active learning can also be integrated with TL for instance selection \cite{drwuSMC2014,drwuTNSRE2016}.

Instance weighting has also been used in deep TL to handle NT. Wang \emph{et al.} \cite{wang2019characterizing} developed a discriminator gate to achieve both adversarial adaptation and class-level weighting of the source samples. They used the output of a discriminator to estimate the distribution density ratio of two domains at each specific feature point:
\begin{align}
\frac{P_t(\bm{x},y)}{P_s(\bm{x},y)}=\frac{D(\bm{x},y)}{1-D(\bm{x},y)},
\end{align}
where $D(\bm{x},y)$ is the output of the discriminator when the input is the concatenation of the feature representation $\bm{x}$ and its predicted label $y$. The supervised learning loss is:
\begin{align}
\begin{split}
\mathcal{L}(C,F) & =\mathbb{E}_{\bm{x}_j,y_j\sim \mathcal{T}_l}[\ell(C(F(\bm{x}_j)),y_j)]\\
& +\lambda\mathbb{E}_{\bm{x}_i,y_i\sim \mathcal{S}}[w(\bm{x}_i,y_i)\ell(C(F(\bm{x}_i)),y_i)],
\end{split}
\end{align}
where $C$ and $F$ are respectively the classifier and the feature extractor, and $w(\bm{x}_i,y_i)={D(\bm{x}_i,y_i)}/({1-D(\bm{x}_i,y_i)})$ is the weight of a source sample $\bm{x}_i$. Wang \emph{et al.} \cite{wang2019characterizing} demonstrated that this approach can remarkably mitigate NT.

\subsubsection{Feature Level Transferability Enhancement}

A popular strategy for feature level transferability enhancement is to learn a common latent feature space, where the features of different domains become more consistent.

Long \emph{et al.} \cite{long2012dual} proposed dual TL to distinguish between the common and domain-specific latent factors automatically. Its main idea is to find a latent feature space that can maximally help the classification in the target domain, formulated as an optimization problem of non-negative matrix tri-factorization:
\begin{align}\label{eq:NMTF}
\min_{U_0, U_S, H, V_S \geq 0} \quad \left \| X_S - [U_0, U_S] H V_S^{\top} \right \|,
\end{align}
where $X_S$ is the source domain feature matrix, $U_0$ and $U_S$ are respectively common feature clusters and domain specific feature clusters, $V_S$ is a sample cluster assignment matrix, and $H$ is the association matrix. (\ref{eq:NMTF}) minimizes the marginal distribution discrepancy between different domains to enable optimal knowledge transfer. Shi \emph{et al.} \cite{shi2013twin} proposed a twin bridge transfer approach, which uses latent factor decomposition of users and similarity graph transfer to facilitate knowledge transfer to reduce NT. This idea was also investigated in \cite{rajesh2017annoyed}, which seeks a latent feature space of the source and target data to minimize the distribution discrepancy.

Another prominent line of work, particularly used in deep TL, is to enhance the feature transferability at the time of feature representations learning. Yosinski \emph{et al.} \cite{yosinski2014tl} defined feature transferability based on its specificity to the domain in which it is trained and its generality.

Multiple approaches have been proposed to compute and enhance the feature transferability \cite{chen2018knowledge,chen2019catastrophic,chen2019transfer, chen2020harmonizing}. For example, Chen \emph{et al.} \cite{chen2019catastrophic} found that features with small singular values have low transferability in deep network fine-tuning; they thus proposed a regularization term to reduce NT, by suppressing the small singular values of the feature matrices.

Unfortunately, focusing only on improving the feature transferability may lead to poor discriminability. It is necessary to consider both feature transferability and discriminability to mitigate NT. To this end, Chen \emph{et al.} \cite{chen2019transfer} proposed to enhance the feature transferability with guaranteed acceptable discriminability by using batch spectral penalization regularization on the largest few singular values.

To reduce the negative impact of noise in the learned feature spaces, Xu \emph{et al.} \cite{xu2015discriminative} introduced a sparse matrix in unsupervised TL to model the feature noise. The loss function with noise minimization is:
\begin{align}
\begin{split}
\min_{P,Z,E} &\ \frac 1 2\phi(P,Y,X_S)+\|Z\|_{*}+\alpha\|Z\|_{1}+\beta\|E\|_1 \\
& \text{s.t.} \ P^{\top}X_t=P^{\top}X_sZ+E,
\end{split}
\end{align}
where $P$, $Z$ and $E$ are the transformation matrix, reconstruction matrix and noise matrix, respectively, $\phi(P,Y,X_S)$ is a discriminant subspace learning function, and $\|\cdot\|_*$ is the nuclear norm of a matrix. The goal is to align the source and target domains in a common low-rank sparse space with noise suppression.

\subsection{Model Transferability Enhancement}

Model transferability enhancement can be achieved through transferable normalization (TransNorm) \cite{wang2019transferable}, adversarial robust training, etc.

TransNorm \cite{wang2019transferable} reduces domain shift in batch normalization \cite{ioffe2015batch}, and is usually used after the convolutional layer to enhance the model transferability. Let the mean and variance of the source domain be $\bm{u}_s$ and $\bm{\sigma}_s$, and the target domain be $\bm{u}_t$ and $\bm{\sigma}_t$. TransNorm quantifies the domain distance as
\begin{align}
\bm{d}^{(j)}=\left\|\frac {\bm{u}_s^{(j)}} {{\bm{\sigma}_s^2}^{(j)}+\epsilon} - \frac {\bm{u}_t^{(j)}} {{\bm{\sigma}_t^2}^{(j)}+\epsilon}\right\|,
\end{align}
where $j$ denotes the $j$-channel in a layer that TransNorm is applied to. Then, it uses distance-based probability $\bm{\alpha}$ to adapt each channel according to its transferability:
\begin{align}
\bm{\alpha}^{(j)}=\frac {c(1+\bm{d}^{(j)})^{-1}} {\sum_{k=1}^c(1+\bm{d}^{(k)})^{-1}}.
\end{align}

Another way to enhance the model transferability is to improve its robustness to adversarial examples \cite{madry2018towards}. Adversarial examples are slightly perturbed inputs aiming to fool a machine learning model \cite{drwuNSR2021,drwuUAP2021}. An adversarially robust model that is resilient to such adversarial examples can be achieved by replacing the standard empirical risk minimization loss with a robust optimization loss \cite{madry2018towards}:
\begin{align}
\min_{\bm{\theta}}\ \mathbb{E}_{(\bm{x},y)\sim D}\left[\max_{\|\bm{\delta}\|_2\le \varepsilon}\ell(\bm{x}+\bm{\delta},y;\bm{\theta})\right],
\end{align}
where $\bm{\delta}$ is a small perturbation, $\varepsilon$ a hyper-parameter to control the perturbation magnitude, and $\bm{\theta}$ the set of model parameters.

Several recent studies found that adversarially robust models have better transferability \cite{liang2020does,salman2020adversarially,deng2021adversarial}. Salman \emph{et al.} \cite{liang2020does} empirically verified that adversarially robust networks achieved higher transfer accuracies than standard ImageNet models, and increasing the width of a robust network may increase its transfer performance gain. Liang \emph{et al.} \cite{salman2020adversarially} found a strong positive correlation between adversarial transferability and knowledge transferability; thus, increasing the adversarial transferability may benefit knowledge transferability.

\subsection{Target Prediction Enhancement}

TL is frequently applied to the target domain with few labeled samples but abundant unlabeled ones. Similar to semi-supervised learning \cite{lee2013pseudo}, pseudo-labels can be used to exploit these unlabeled samples in TL. Soft pseudo-labeling, selective pseudo-labeling and cluster enhanced pseudo-labeling could be used to mitigate NT.

Soft pseudo-labeling assigns each unlabeled sample to different classes with different probabilities, rather than a single class, in order to alleviate label noise from a weak source classifier \cite{wang2020spl}. For example, in multi-adversarial DA \cite{pei2018multi}, the soft pseudo-label of a target sample is used to indicate how much this sample should be emphasized by different class-specific domain discriminators. In deep unsupervised DA, Ge \emph{et al.} \cite{ge2020mutual} introduced a soft softmax-triplet loss based on the soft pseudo-labels, which outperformed hard labeling.

Selective pseudo-labeling is another strategy to enhance target prediction. Its main motivation is to select the unlabeled samples with high confidence as the training targets. For instance, Gui \emph{et al.} \cite{gui2018negative} developed an approach to predict when NT would occur. They identified and removed the noisy samples in the target domain to reduce class noise accumulation in future training iterations. Wang and Breckon \cite{wang2020spl} proposed selective pseudo-labeling to progressively select a subset containing $mn_t/T$ high-probability target samples in the $m$-th iteration, where $T$ is the number of iterations of the learning process. Their experiments showed that this simple strategy generally improved the target prediction performance.

Cluster enhanced pseudo-labeling is based on soft pseudo-labeling, but further explores the unsupervised clustering information to enhance target domain prediction \cite{liang2019exploring,wang2020spl,liang2020shot}. For example, Liang \emph{et al.} \cite{liang2020shot} developed a self-supervised pseudo-labeling approach to alleviate harmful effects resulted from inaccurate adaptation network outputs. Its main idea is to perform weighted $k$-means clustering on the target data to get the class means,
\begin{align}
\begin{split}
\mu_k = \frac {\sum_{x_t\in \mathcal{T}}\delta_k(\hat{\theta}_t(x_t))\hat{g}_t(x_t)} {\sum_{x_t\in \mathcal{T}}\delta_k(\hat{\theta}_t(x_t))},
\end{split}
\end{align}
where $\hat{\theta}_t = \hat{f}_t(\hat{g}_t(x_t))$ denotes the learned target network parameters, $\hat{g}_t(\cdot)$ is a feature extractor, $\hat{f}_t(\cdot)$ is a classification layer, and $\delta_k(\cdot)$ denotes the $k$-th element in the soft-max output. With the learned robust feature centroids $\mu_k$, the pseudo-labels can be updated by a nearest centroid classifier:
\begin{align}
\begin{split}
\hat{y}_t = \arg \min_k \ \|\hat{g}_t(x_t) - \mu_k\|^2_2.
\end{split}
\end{align}
The centroids and pseudo-labels can be optimized iteratively to obtain better pseudo-labels.

\section{NT in Related Fields}

NT has also been found and studied in several related fields, including multi-task learning \cite{ruder2017overview}, lifelong learning \cite{parisi2019continual}, and adversarial attacks \cite{madry2018towards}.

\subsection{Multi-Task Learning}

Multi-task learning solves multiple learning tasks jointly, by exploiting commonalities and differences across them. Similar to TL, it needs to facilitate positive transfer among tasks to improve the overall learning performance on all tasks. Previous studies \cite{yu2020gradient,wang2020gradient} have observed that conflicting gradients among different tasks may induce NT (also known as negative interference). Various techniques have been explored to remedy negative interference, such as altering the gradients directly \cite{chen2018gradnorm,kendall2018multi}, weighting tasks \cite{liu2019loss}, learning task relatedness \cite{zhang2012convex,shui2019amtnn}, routing networks \cite{rusu2016progressive,rosenbaum2019routing}, and searching for Pareto solutions \cite{sener2018multi,lin2019pareto}, etc.

As a concrete example of multi-task learning, multilingual models have demonstrated success in processing tens or even hundreds of languages simultaneously \cite{devlin2018bert,conneau2019unsupervised,arivazhagan2019massively}.
However, not all languages can benefit from this training paradigm. Studies \cite{wang2020negative} have revealed NT in multilingual models, especially for high-resource languages \cite{arivazhagan2019massively}.
Possible remedies include parameter soft-sharing \cite{guo2018multi}, meta-learning \cite{wang2020negative}, and gradient vaccine \cite{wang2020gradient}.

\subsection{Lifelong Learning}

Lifelong learning learns a series of tasks in a sequential order, without revisiting previously seen data.
While the goal is to master all tasks in a single model, there are two key challenges, which may lead to NT. First, the model may forget earlier knowledge when trained on new tasks, known as catastrophic forgetting \cite{mccloskey1989catastrophic,doan2021theoretical}. Second, transferring from early tasks may hurt the performance in later tasks. Existing literature mainly studies how to mitigate catastrophic forgetting using regularization \cite{chaudhry2018efficient,sprechmann2018memory}, memory replay \cite{lopez2017gradient,rolnick2019experience,d2019episodic}, parameter isolation \cite{Mallya2018,Serra2018}, etc., whereas forward NT in lifelong learning is less investigated \cite{wang2020efficient}.

\subsection{Adversarial Attacks}

Adversarial attack aims at learning perturbations on the training data or models, and then sabotaging the test performances. Researchers found that adversarial examples have high transferability, and the unsecured source data or models highly affect the target learning performance. As a result, the target model performs poorly, and NT may occur. For example, the white-box teacher model, black-box student model and TL parameters can be affected by evasion attacks \cite{wang2018great,cheng2019improving,abdelkader2020headless}, and source data by backdoor attacks \cite{wang2020backdoor,drwuEngineering2021,drwuAP2021}.

\section{Method Comparison}

According to the scheme shown in Fig.~\ref{fig:overview}, we have introduced secure transfer, domain similarity estimation, distant transfer, and tens of NT mitigation approaches. To see the forest for the trees, we compare their characteristics and differences in Table~\ref{tab:comparison}.

\begin{table*}[htbp]\centering \renewcommand\arraystretch{1.3}
  \caption{Comparison of approaches for mitigating NT. ``TL Category" groups all methods into five categories \cite{zhang2019tdl}: instance adaptation, feature adaptation, model adaptation, deep TL, and adversarial TL. ``Ability to overcome NT" has three levels: guaranteed ($\star\star\star$), very probable ($\star\star$), and possible ($\star$). ``Factors of NT" includes four elements mentioned in Section~\ref{sect:factor}: domain divergence (D), transfer algorithm (A), source data quality (S), and target data quality (T).}
    \begin{tabular}{c|c|L{2.6cm}p{4.3cm}p{3cm}p{2cm}}  \hline
    \multicolumn{2}{c|}{Strategy} & [Reference] Approach & TL Category & Ability to Overcome NT & Factors of NT \\  \hline
    \multicolumn{2}{c|}{\multirow{14}{1.8cm}{\centering Domain Similarity Estimation}}
        & \cite{gretton2012kernel} MMD     & Feature adaptation     & $\star\star$     & D, S \\
     \multicolumn{1}{c}{} &  & \cite{Gong2012} ROD     & Feature adaptation     & $\star\star$     & S \\
     \multicolumn{1}{c}{} &  & \cite{lin2017improving} cTL     & Feature adaptation     & $\star\star$     & D, A, S \\
     \multicolumn{1}{c}{}&   & \cite{zhang2020manifold} DTE     & Feature adaptation     & $\star\star$     & S \\
     \multicolumn{1}{c}{} &  & \cite{azab2019weighted} WTL     & Model adaptation     & $\star\star$     & D, A, S \\
     \multicolumn{1}{c}{}  & & \cite{yao2010boosting}     & Instance adaptation     & $\star\star$     & D, S \\
     \multicolumn{1}{c}{} &  & \cite{xie2017selective} Q-statistic    & Model adaptation     & $\star\star$     & D, S \\
     \multicolumn{1}{c}{} &  & \cite{ben2007analysis} $\mathcal{A}$-distance     & Instance adaptation     & $\star\star$     & D, S \\
     \multicolumn{1}{c}{} &  & \cite{xu2018deep} DCTN     & Deep TL/Feature adaptation     & $\star\star$     & D, A, S \\
     \multicolumn{1}{c}{} &  & \cite{wu2020continuous} TransLATE     & Adversarial TL     & $\star\star$     & D, A, S, T \\
     \multicolumn{1}{c}{}&   & \cite{Meiseles2020} MSC     & Deep TL     & $\star\star$     & S \\
     \multicolumn{1}{c}{} &  & \cite{tran2019transfer} NCE     & Deep TL/Model adaptation     & $\star\star$     & A, S \\
     \multicolumn{1}{c}{}&   & \cite{afridi2018automated}     & Deep TL/Model adaptation     & $\star\star$     & A, S \\
     \multicolumn{1}{c}{} &  & \cite{leep2020} LEEP     & Deep TL/Model adaptation     & $\star\star$     & A, S \\
     \multicolumn{1}{c}{} &  & \cite{huang2021frustratingly} TransRate  & Deep TL/Model adaptation  & $\star\star$   & A, S \\  \hline
    \multicolumn{2}{c|}{\multirow{6}{*}{\centering Secure Transfer}}
        & \cite{cao2010adaptive} AT-GP    & Model adaptation     & $\star\star\star$     & A, T \\
      \multicolumn{1}{c}{}&  & \cite{abdullah2018deep}     & Deep TL     & $\star\star\star$     & A, T \\
      \multicolumn{1}{c}{}&  & \cite{li2019towards} SAFEW     & Model adaptation     & $\star\star\star$     & A, T \\
      \multicolumn{1}{c}{}&  & \cite{kuzborskij2013stability}     & Model adaptation     & $\star\star\star$     & A, T \\
      \multicolumn{1}{c}{}&  & \cite{yoon2018novel} PTL     & Model adaptation     & $\star\star\star$     & A, T \\
      \multicolumn{1}{c}{}&  & \cite{Sorocky2021}     & Model adaptation     & $\star\star\star$     & A, T \\  \hline
    \multicolumn{2}{c|}{\multirow{4}{*}{\centering Distant Transfer}}
        & \cite{tan2015transitive} TTL     & Instance adaptation     & $\star$     & D, A, S  \\
     \multicolumn{1}{c}{}&   & \cite{tan2017distant} DDTL     & Instance adaptation     & $\star$     & D, A, S \\
     \multicolumn{1}{c}{}&   & \cite{niu2020distant} DFF     & Deep TL/Feature adaptation     & $\star$     & D, A, S \\
     \multicolumn{1}{c}{}&   & \cite{Xie2016transfer}     & Feature adaptation     & $\star$     & A, S \\  \hline
    \multirow{25}{1.5cm}{\centering NT Mitigation}
    &\multirow{14}{1.8cm}{\centering Data\\ Transferability Enhancement}
        & \cite{wang2018towards} PW-MSTL     & Instance adaptation     & $\star\star$     & A, S \\
     &   & \cite{zuo2021attention} ABMSDA    & Deep TL/Feature adaptation     & $\star\star$     & D, A, S \\
     &   & \cite{ahmed2021unsupervised} DECISION    & Deep TL/Model adaptation   & $\star\star$    & D, A, S, T \\
      &  & \cite{seah2012combating} PDM     & Instance adaptation     & $\star\star$     & A, S \\
      &  & \cite{drwuTNSRE2016} AwAR     & Feature adaptation    & $\star\star$     & D, A, T \\
      &  & \cite{xu2020multi} MCTML    & Instance adaptation     & $\star\star$     & D, A, S \\
      &  & \cite{peng20atl} ATL     & Feature adaptation     & $\star\star$     & D, A, S \\
      &  & \cite{wang2019characterizing} GATE     & Adversarial TL     & $\star\star$     & D, A, S \\
      &  & \cite{long2012dual} DTL     & Feature adaptation     & $\star\star$     & D, A, S \\
      &  & \cite{shi2013twin} TBT     & Feature adaptation     & $\star\star$     & D, A, S \\
      &  & \cite{xu2015discriminative}     & Feature adaptation     & $\star\star$     & D, A, S \\
      &  & \cite{rajesh2017annoyed} DTL     & Feature adaptation     & $\star\star$     & D, A, S \\
      &  & \cite{chen2019catastrophic} BSS     & Deep TL/Model adaptation     & $\star\star$     & A, S \\
      &  & \cite{chen2019transfer} BSP     & Adversarial TL     & $\star\star$     & D, A, S \\
      &  & \cite{chen2020harmonizing} HTCN     & Adversarial TL     & $\star\star$     & D, A, S \\ \cline{2-6}
    &\multirow{4}{1.8cm}{\centering Model\\ Transferability Enhancement}
        & \cite{wang2019transferable} TransNorm     & Deep TL     & $\star\star$     & A \\
      &  & \cite{liang2020does} AT     & Deep TL     & $\star\star$     & A \\
      &  & \cite{salman2020adversarially}     & Deep TL     & $\star\star$     & A \\
      &  & \cite{deng2021adversarial}     & Deep TL     & $\star\star$     & A \\  \cline{2-6}
    &\multirow{7}{1.8cm}{\centering Target Prediction Enhancement}
        & \cite{pei2018multi} MADA     & Adversarial TL     & $\star\star$     & D, A, T \\
      &  & \cite{ge2020mutual} MMT    & Deep TL/Model adaptation     & $\star\star$     & D, A, T \\
      &  & \cite{gui2018negative} NTD     & Instance adaptation     & $\star\star$     & D, A, T \\
      &  & \cite{wang2020spl} SPL     & Feature adaptation     & $\star\star$     & D, A, T \\
      &  & \cite{liang2019exploring} PACET     & Feature adaptation     & $\star\star$     & D, A, T \\
      &  & \cite{liang2020shot} SHOT     & Deep TL/Model adaptation     & $\star\star$     & D, A, T \\ \hline
    \end{tabular}
  \label{tab:comparison}
\end{table*}

Several observations can be made from Table~\ref{tab:comparison}:
\begin{enumerate}
\item Most current works in the NT literature focus on NT mitigation and domain similarity estimation.
\item NT mitigation research mainly focuses on data transferability enhancement.
\item The research on overcoming NT spreads across many different categories of TL, indicating that it has received extensive attention in TL.
\item Most secure transfer strategies are based on model adaptation.
\item Most NT mitigation approaches consider two or more factors of NT (domain divergence, transfer algorithm, source data quality, and target data quality).
\end{enumerate}

\section{Conclusions and Future Research}

TL utilizes data or knowledge from one or more source domains to facilitate the learning in a target domain, which is particularly useful when the target domain has very few or no labeled data. NT is undesirable in TL, and has been attracting increasing research interest recently. This paper has systematically categorized and reviewed about fifty representative approaches for handling NT, from four perspectives: secure transfer, domain similarity estimation, distant transfer, and NT mitigation. Besides, some fundamental concepts, e.g., the definition of NT, the factors of NT, and related fields of NT, are also introduced. To our knowledge, this is the first comprehensive survey on NT.

We suggest the following guidelines in coping with NT:
\begin{itemize}
\item If the domain similarity can be estimated, then we can choose different strategies according to the similarity: directly concatenate the source and target domain data and train a machine learning model if the similarity is high; choose a NT mitigation approach if the similarity is medium; or, use distant transfer or no transfer at all if the similarity is low.
\item If the domain similarity cannot be estimated, then secure transfer may be used.
\end{itemize}

The following directions may be considered in future research:
\begin{enumerate}
\item Develop secure transfer approaches that correspond to popular TL paradigms such as unsupervised DA, few shot TL, and adversarial TL.
\item Investigate NT mitigation approaches for regression problems. Currently most such approaches are for classification problems.
\item Ensure positive transfer in challenging open environments, which may include continual data stream, heterogeneous features, private sources, unclear domain boundaries, unseen/unknown categories, etc.
\item Design theoretical tests or empirical procedures to identify the exact factors leading to NT in a specific application, which can help us choose the most appropriate approach to mitigate NT.
\end{enumerate}

\section*{Acknowledgment}

This work was supported by the Hubei Province Funds for Distinguished Young Scholars under Grant 2020CFA050, the Technology Innovation Project of Hubei Province of China under Grant 2019AEA171, the National Natural Science Foundation of China under Grants 61873321 and U1913207, and the International Science and Technology Cooperation Program of China under Grant 2017YFE0128300. The authors would also like to thank Mr. Zirui Wang of the Carnegie Mellon University for insightful discussions.

\bibliographystyle{IEEEtran} \bibliography{NTL_survey_v1}

\end{document}